\documentclass[10pt,twocolumn,letterpaper]{article}

\usepackage{cvpr}
\usepackage{times}
\usepackage{epsfig}
\usepackage{graphicx}
\usepackage{amsmath}
\usepackage{amssymb}
\usepackage{amsthm}
\usepackage{soul}
\usepackage{subcaption}

\renewcommand{\eg}{\textit{e.g.}}
\renewcommand{\ie}{\textit{i.e.}}
\newcommand{\set}[1]{\{{#1}\}}
\newcommand{\D}{\mathcal{D}}
\newcommand{\KL}[2]{\mathop{KL}({\textstyle #1}\,\|\,{{\textstyle #2}})}

\def\tr{\mathsf{tr}}

\newcommand{\N}{\mathcal{N}}
\newcommand{\E}{\mathbb{E}}
\newcommand{\DDX}[1]{\frac{\partial}{\partial{#1}}}
\newcommand{\DXDY}[2]{\frac{\partial{#1}}{\partial{#2}}}
\newcommand{\HXY}[2]{\frac{\partial^2\ell}{\partial{#1}\partial{#2}}}
\newcommand{\FXY}[2]{\frac{\partial \ell}{\partial{#1}}\left(\frac{\partial \ell}{\partial{#2}}\right)^T}
\newcommand{\ttv}{\textsc{task2vec} }
\newcommand{\mtv}{\textsc{model2vec}}

\makeatletter
\renewcommand{\paragraph}{ \@startsection{paragraph}{4} {\z@}{1.25ex \@plus 1ex \@minus .2ex}{-1em} {\normalfont\normalsize\bfseries} }
\makeatother

\usepackage[pagebackref=true,breaklinks=true,colorlinks,bookmarks=false]{hyperref}
\usepackage{cleveref}
 \cvprfinalcopy %
\def\cvprPaperID{2796} %

\ifcvprfinal\fi
\begin{document}

\title{\textsc{Task2Vec:}  Task Embedding for Meta-Learning}

\author{Alessandro Achille \\ UCLA and AWS \\
{\tt \footnotesize achille@cs.ucla.edu} \and Michael Lam \\ AWS \\
{\tt\footnotesize michlam@amazon.com} \and Rahul Tewari \\AWS \\
{\tt\footnotesize tewarir@amazon.com} \and Avinash Ravichandran \\ AWS \\
{\tt\footnotesize ravinash@amazon.com} \and Subhransu Maji \\ UMass  and AWS \\
{\tt\footnotesize smmaji@amazon.com} \and Charless Fowlkes \\ UCI and AWS \\
{\tt\footnotesize fowlkec@amazon.com} \and Stefano Soatto \\ UCLA and AWS \\
{\tt\footnotesize soattos@amazon.com} \and Pietro Perona
\\
Caltech and AWS\\
{\tt\footnotesize peronapp@amazon.com}
}

\maketitle

\begin{abstract}
We introduce a method to provide vectorial representations of visual classification tasks which can be used to reason about the nature of those tasks and their relations. Given a dataset with ground-truth labels and a loss function defined over those labels, we process images through a ``probe network'' and compute an embedding based on estimates of the Fisher information matrix associated with the probe network parameters.  This provides a fixed-dimensional embedding of the task that is independent of details such as the number of classes and does not require any understanding of the class label semantics.
We demonstrate that this embedding is capable of predicting task similarities that match our intuition about semantic and taxonomic relations between different visual tasks ({\em e.g.}, tasks based on classifying different types of plants are similar). We also demonstrate the practical value of this framework for the meta-task of selecting a pre-trained feature extractor for a new task. We present a simple meta-learning framework for learning a metric on embeddings that is capable of predicting which feature extractors will perform well. Selecting a feature extractor with task embedding obtains a performance close to the best available feature extractor, while costing substantially less than exhaustively training and evaluating on all available feature extractors.
\end{abstract}

\begin{figure*}[htb!]
    \centering
    \includegraphics[width=.8\linewidth]{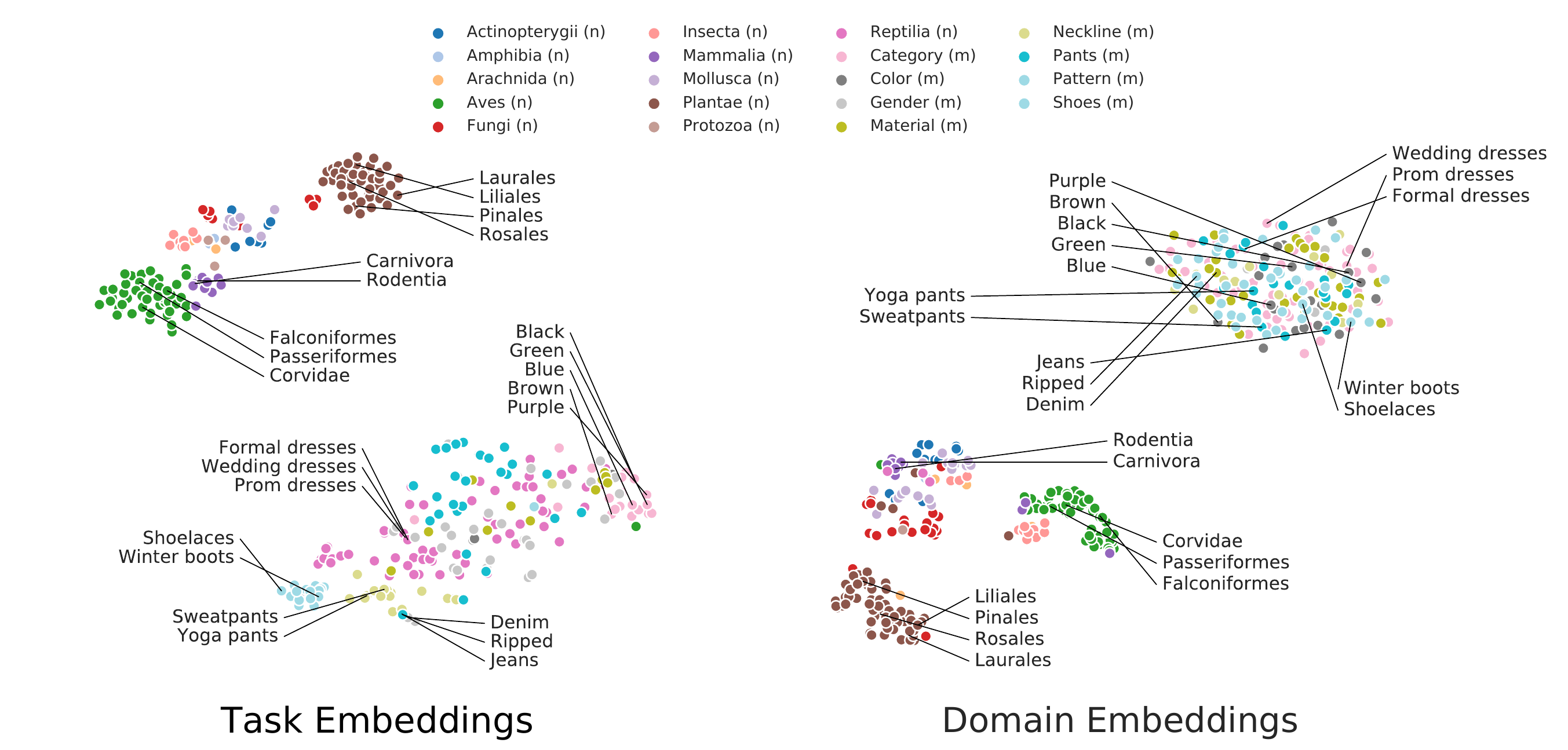}
    \caption{
    \textbf{Task embedding across a large library of tasks} (best seen magnified). \textbf{(Left)} T-SNE visualization of the embedding of tasks extracted from the iNaturalist, CUB-200, iMaterialist datasets. Colors indicate ground-truth  grouping of tasks based on taxonomic or semantic types.
    Notice that the bird classification tasks extracted from CUB-200 embed near the bird classification task from iNaturalist, even though the original datasets are different. iMaterialist is well separated from iNaturalist, as it entails very different tasks (clothing attributes). Notice that some tasks of similar type (such as color attributes) cluster together but attributes of different task types may also mix when the underlying visual semantics are correlated. For example, the tasks of jeans (clothing type), denim (material) and ripped (style) recognition are close in the task embedding. \textbf{(Right)} T-SNE visualization of the domain embeddings (using mean feature activations) for the same tasks. Domain embedding can distinguish
    iNaturalist tasks from iMaterialist tasks due to differences in the two problem
    domains. However, the fashion attribute tasks on iMaterialist all share the same domain and only differ in their labels. In this case, the domain embeddings collapse to a region without recovering any sensible structure.}
    \label{fig:taxonomic_qualitative}
\end{figure*}

\section{Introduction}

The success of Deep Learning hinges in part on the fact that models learned for one task can be used on other related tasks. Yet, no general framework exists to describe and learn relations between tasks. We introduce the \textsc{task2vec} embedding, a technique to represent tasks as elements of a vector space based on the Fisher Information Matrix. The norm of the embedding correlates with the complexity of the task, while the distance between embeddings captures semantic similarities between tasks (Fig.~\ref{fig:taxonomic_qualitative}). When other natural distances are available, such as the taxonomical distance in biological classification, we find that the embedding distance correlates positively with it (Fig.~\ref{fig:taxonomical_distance}). Moreover, we introduce an asymmetric distance on tasks which correlates with the transferability between tasks.

Computation of the embedding leverages a duality between network parameters (weights) and outputs (activations) in a deep neural network (DNN): Just as the activations of a DNN trained on a complex visual recognition task are a rich representation of the input images, we show that the gradients of the weights relative to a task-specific loss are a rich representation of the task itself. Specifically, given a task defined by a dataset $\D=\set{(x_i,y_i)}_{i=1}^N$ of labeled samples, we feed the data through a pre-trained reference convolutional neural network which we call ``\textit{probe network}'', and compute the diagonal Fisher Information Matrix (FIM) of the network filter parameters to capture the structure of the task (Sect.~\ref{sec:fisher}).  Since the architecture and weights of the probe network are fixed, the FIM provides a fixed-dimensional representation of the task. We show this embedding encodes the ``difficulty'' of the task, characteristics of the input domain, and which features of the probe network are useful to solve it (Sect.~\ref{sec:task2vec}).

Our task embedding can be used to reason about the space of tasks and solve meta-tasks. As a motivating example, we study the problem of selecting the best pre-trained feature extractor to solve a new task. This can be particularly valuable when there is insufficient data to train or fine-tune a generic model, and transfer of knowledge is essential.
\ttv depends solely on the task, and ignores interactions with the model which may however play an important role. To address this, we learn a joint task and model embedding, called \mtv, in such a way that models whose embeddings are close to a task exhibit good perfmormance on the task. We use this to select an expert from a given collection, improving  performance relative to fine-tuning a generic model trained on ImageNet and obtaining close to ground-truth optimal selection. We discuss our contribution in relation to prior literature in Sect.~\ref{sec:related-work}, after presenting our empirical results in Sect.~\ref{sec:experiments}.

\section{Task Embeddings via Fisher Information}

\label{sec:fisher}

Given an observed input $x$ (\eg, an image) and an hidden task variable $y$ (\eg, a label), a deep network is a family of functions $p_w(y|x)$ parametrized by weights $w$, trained to approximate the posterior $p(y|x)$ by minimizing the (possibly regularized) cross entropy loss $H_{p_w,\hat{p}}(y|x) = \E_{x,y \sim \hat{p}}[-\log p_w(y|x)]$, where $\hat{p}$ is the empirical distribution defined by the training set $\D=\set{(x_i,y_i)}_{i=1}^N$.
It is useful, especially in transfer learning, to think of the network as composed
of two parts: a feature extractor which computes some representation $z=\phi_w(x)$ of the input data, and a ``head,'' or classifier, which encodes the distribution $p(y|z)$ given the representation $z$.

Not all network weights are equally useful in predicting the task variable: the importance, or ``informative content,'' of a weight for the task can be quantified by considering a perturbation $w'=w + \delta w$ of the weights, and measuring the average Kullbach-Leibler (KL) divergence between the original output distribution $p_{w}(y|x)$ and the perturbed one $p_{w'}(y|x)$. To second-order approximation, this is
\[
\E_{x\sim \hat{p}} \KL{p_{w'}(y|x)}{p_w(y|x)} = \delta w \cdot F \delta w + o(\delta w^2),
\]
where $F$ is the Fisher information matrix (FIM):
\[
F = \E_{x, y \sim \hat{p}(x) p_w (y|x)} \left[ \nabla_w \log p_w(y|x) \nabla_w \log p_w(y|x)^T \right].
\]
that is, the expected covariance of the scores (gradients of the log-likelihood) with respect to the model parameters.

The FIM is a Riemannian metric on the space of probability distributions \cite{amari2000methods}, and provides a measure of the information a particular parameter (weight or feature) contains about the joint distribution $p_w(x,y)=\hat{p}(x)p_w(y|x)$: If the classification performance for a given task does not depend strongly a parameter, the corresponding entry in the FIM will be small. The FIM is also related to the (Kolmogorov) complexity of a task, a property that can be used to define a computable metric of the learning distance between tasks \cite{achille2018kolmogorov}. Finally, the FIM can be interpreted as an easy-to-compute positive semidefinite upper-bound to the Hessian of the cross-entropy loss, and coincides with it at local minima \cite{martens14new}. In particular, ``flat minima'' correspond to weights that have, on average, low (Fisher) information \cite{achille2018emergence,hochreiter1997flat}.

\subsection{\ttv embedding using a probe network}
\label{sec:task2vec}
While the network activations capture the information in the input image which are needed to infer the image label, the FIM indicates the set of feature maps which are more informative for solving the current task. Following this intuition, we use the FIM to represent the task itself. However, the FIMs computed on different networks are not directly comparable.  To address this, we use single ``probe'' network pre-trained on ImageNet
as a feature extractor and re-train only the classifier layer on any given task, which usually can be done efficiently. After training is complete, we compute the FIM for the feature extractor parameters.

Since the full FIM is unmanageably large for rich probe networks based on CNNs, we make two additional approximations. First, we only consider the diagonal entries, which implicitly assumes that correlations between different filters in the probe network are not important.  Second, since the weights in each filter are usually not independent, we average the Fisher Information for all weights in the same filter. The resulting representation thus has fixed size, equal to the number of filters in the probe network. We call this embedding method \textsc{task2vec}.

\paragraph{Robust Fisher computation}
Since the FIM is a local quantity, it is  affected by the local geometry of the training loss landscape, which is highly irregular in many deep network architectures \cite{li2017visualizing}, and may be too noisy when trained with few samples. To avoid this problem, instead of a direct computation, we use a more robust estimator that leverages connections to variational inference. Assume we perturb the weights $\hat{w}$ of the network with Gaussian noise $\N(0, \Lambda)$ with precision matrix $\Lambda$, and we want to find the optimal $\Lambda$ which yields a good expected error, while remaining close to an isotropic prior $\N(\hat{w}, \lambda^2 I)$. That is, we want to find $\Lambda$ that minimizes:
\begin{multline*}
L(\hat{w}; \Lambda) = \E_{w\sim \N(\hat{w},\Lambda)} [H_{p_w,\hat{p}}p(y|x)] \\
+ \beta \KL{\N(0, \Lambda)}{\N(0, \lambda^2 I)},
\end{multline*}
where $H$ is the cross-entropy loss and $\beta$ controls the weight of the prior. Notice that for $\beta=1$ this reduces to the Evidence Lower-Bound (ELBO) commonly used in variational inference. Approximating to the second order, the optimal value of $\Lambda$ satisfies (see Supplementary Material):
\[
\frac{\beta}{2N} \Lambda = F + \frac{\beta\lambda^2}{2N} I.
\]
Therefore, $\frac{\beta}{2N} \Lambda \sim F + o(1)$ can be considered as an estimator of the FIM $F$, biased towards the prior $\lambda^2 I$ in the low-data regime instead of being degenerate. In case the task is trivial (the loss is constant or there are too few samples) the embedding will coincide with the prior $\lambda^2 I$, which we will refer to as the \textbf{trivial embedding}. This estimator has the advantage of being easy to compute by directly minimizing the loss  $L(\hat{w}; \Sigma)$ through Stochastic Gradient Variational Bayes \cite{kingma2015variational}, while being  less sensitive to irregularities of the loss landscape than direct computation, since the value of the loss depends on the cross-entropy in a neighborhood of $\hat{w}$ of size $\Lambda^{-1}$. As in the standard Fisher computation, we estimate one parameter per filter, rather than per weight, which in practice means that we constrain $\Lambda_{ii} = \Lambda_{jj}$ whenever $w_i$ and $w_j$ belongs to the same filter. In this case, optimization of $L(\hat{w}; \Lambda)$ can be done efficiently using the local reparametrization trick of \cite{kingma2015variational}.

\begin{figure*}[ht]
    \centering
    \raisebox{0.0\height}{\includegraphics[width=.4\linewidth,trim={0 0 0 4cm},clip]{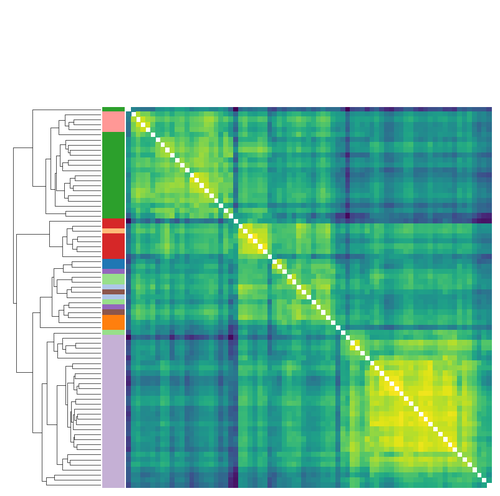}}
    \hspace{0.1in}
    \raisebox{0.12\height}{\includegraphics[width=0.25\linewidth]{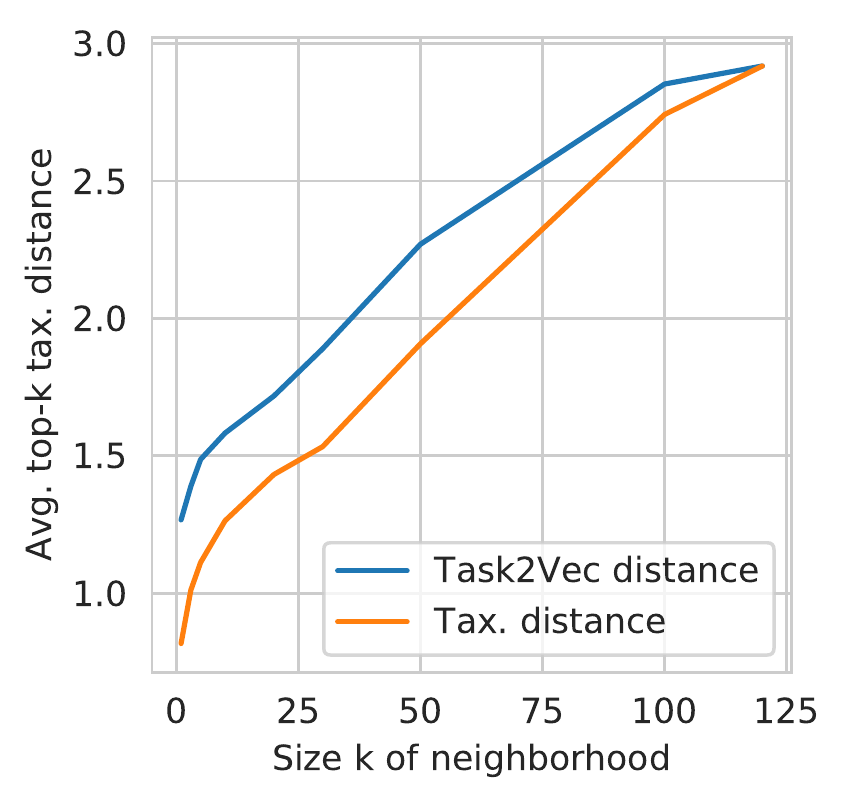}}
    \hspace{-0.05in}
    \raisebox{0.12\height}{\includegraphics[width=0.25\linewidth]{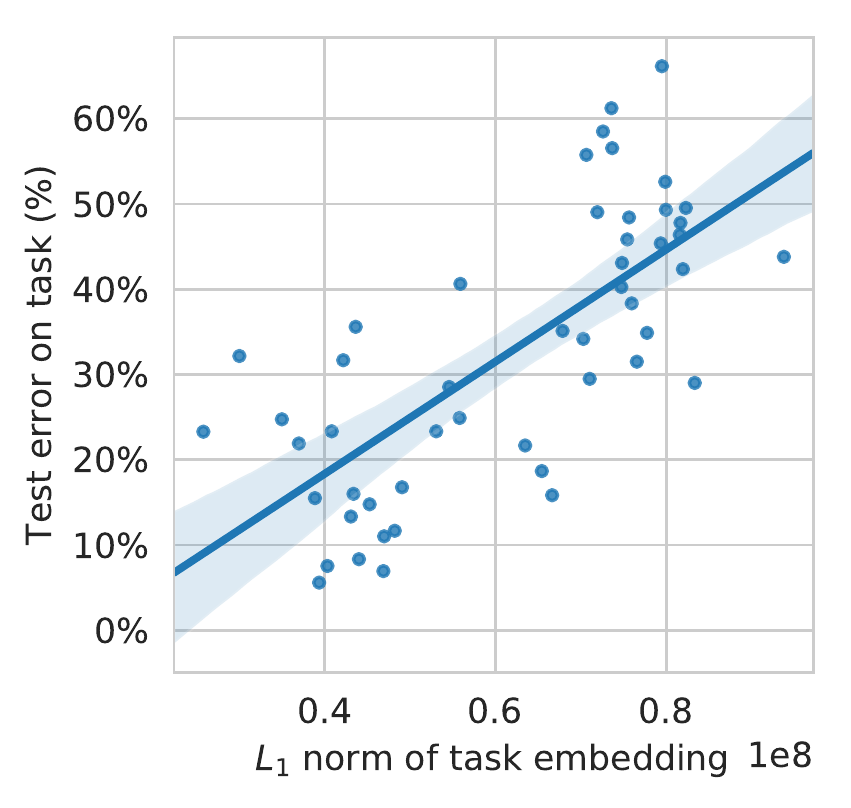}}
     \caption{
     \textbf{Distance between species classification tasks. }
     \textbf{(Left)} Task similarity matrix ordered by hierarchical clustering.  Note that the dendrogram produced by the task similarity matches the taxonomic clusters (indicated by color bar).
     \textbf{(Center)} For tasks extracted from iNaturalist and CUB, we compare the cosine distance between tasks to their taxonomical distance.  As the size of the task embedding neighborhood increases (measured by number of tasks in the neighborhood), we plot the average taxonomical distance of tasks from the neighborhood center. While the task distance does not perfectly match the taxonomical distance (whose curve is shown in orange), it shows a good correlation. Difference are both due to the fact that taxonomically close species may need very different features to be classified, creating a mismatch between the two notions of distance, and because for some tasks in iNaturalist too few samples are provided to compute a good embedding. \textbf{(Right)} Correlation between $L_1$ norm of the task embedding (distance from origin) and test error obtained on the task.
    }
    \label{fig:taxonomical_distance}
\end{figure*}

\subsection{Properties of the \ttv embedding}

The task embedding we just defined has a number of useful properties. For illustrative purposes, consider a two-layer sigmoidal network for which an analytic expression can be derived (see Supplementary Materials). The FIM of the feature extractor parameters can be written using the Kronecker product as
\[
F = \E_{x,y\sim \hat{p}(x)p_w(y|x)} [(y-p)^2 \cdot S \otimes xx^T]
\]
where $p = p_w(y = 1 | x)$ and the matrix $S =  ww^T \odot zz^T \odot (1-z)(1-z)^T$ is an element-wise product of classifier weights
$w$ and first layer feature activations $z$.
It is informative to compare this expression to an embedding based only on the
dataset domain statistics, such as the (non-centered) covariance $C_0 = \E \left[ xx^T\right]$ of the
input data or the covariance $C_1 = \E \left[ zz^T\right]$ of the feature activations.
One could take such statistics as a representative {\em domain embedding} since
they only depend on the marginal distribution $p(x)$ in contrast to the FIM
{\em task embedding}, which depends on the joint distribution $p(x,y)$. These simple expressions highlight some important (and more general) properties of the Fisher embedding
we now describe.

\textbf{Invariance to the label space:} The task embedding does not directly
depend on the task labels, but only on the predicted distribution $ p_w(y| x)$ of the trained model.
Information about the ground-truth labels $y$ is encoded in the weights $w$ which are
a sufficient statistic of the task \cite{achille2018emergence}. In particular, the task embedding is invariant to
permutations of the labels $y$, and has fixed dimension (number of filters of the feature extractor) regardless of the output
space (e.g., k-way classification with varying k).

\textbf{Encoding task difficulty:} As we can see from the expressions above,
if the fit model is very confident in its predictions, $\E[(y-p)^2]$ goes to
zero. Hence, the norm of the task embedding $\|F\|_\star$ scales with the difficulty of
the task for a given feature extractor $\phi$. Figure~\ref{fig:taxonomical_distance} (Right)
shows that even for more complex models trained on real data, the FIM norm correlates with test performance.

\textbf{Encoding task domain:} Data points $x$ that are classified
with high confidence, i.e., $p$ is close to 0 or 1, will have a lower contribution
to the task embedding than points near the decision boundary since $p(1-p)$ is
maximized at $p=1/2$. Compare this to the covariance matrix of the data,
$C_0$, to which all data points contribute equally. Instead, in \ttv
information on the domain is based on data near the decision boundary (task-weighted domain embedding).

\textbf{Encoding useful features for the task:}
The FIM depends on the curvature of the loss function with the diagonal entries capturing
the sensitivity of the loss to model parameters. Specifically, in the two-layer model one
can see that, if a given feature is uncorrelated with $y$, the corresponding
blocks of $F$ are zero. In contrast, a domain embedding based on feature activations
of the probe network (e.g., $C_1$) only reflects which features vary over the dataset
without indication of whether they are relevant to the task.

\section{Similarity Measures on the Space of Tasks}

What metric should be used on the space of tasks? This depends critically on the meta-task we are considering.
As a motivation, we concentrate on the meta-task of selecting the pre-trained feature extractor from a set in order to obtain the best performance on a new training task. There are several natural metrics that may be considered for this meta-task. In this work, we mainly consider:

\paragraph{Taxonomic distance} For some tasks, there is a natural notion of semantic similarity,
for instance defined by sets of categories organized in a taxonomic hierarchy where each task is classification inside a subtree of the hierarchy (\textit{e.g.}, we may say that classifying breeds of dogs is closer to classification of cats than it is to classification of species of plants). In this setting, we can
define
\[
  D_\text{tax}(t_a,t_b) = \min_{i\in S_a, j\in S_b} d(i,j),
\]
where $S_a,S_b$ are the sets of categories in task $t_a,t_b$ and $d(i,j)$ is an
ultrametric or graph distance in the taxonomy tree. Notice that this is a proper distance, and in particular it is symmetric.

\paragraph{Transfer distance.} We define the transfer (or fine-tuning) gain from a task $t_a$ to a task $t_b$ (which we improperly call distance, but is not necessarily symmetric or positive) as the difference in expected performance between a model trained for task $t_b$  from a fixed initialization (random or pre-trained),
and the performance of a model fine-tuned for task $t_b$ starting from a solution of task $t_a$:
\[
  D_{\text{ft}}(t_a \to t_b) = \frac{\E[\ell_{a\to b}] - \E[\ell_b]}{\E[\ell_b]},
\]
where the expectations are taken over all trainings with the selected architecture,  training procedure and network initialization, $\ell_{b}$ is the final test error obtained by training on task $b$ from the chosen initialization, and $\ell_{a \to b}$ is the error obtained instead when starting from a solution to task $a$ and then fine-tuning (with the selected procedure) on task $t_b$.

\subsection{Symmetric and asymmetric \ttv metrics}

By construction, the Fisher embedding on which \ttv is based captures fundamental information about the structure of the task. We may therefore expect that the distance between two embeddings correlate positively with natural metrics on the space of tasks. However, there are two problems in using the Euclidean distance between embeddings: the parameters of the network have different scales, and the norm of the embedding is affected by complexity of the task and the number of samples used to compute the embedding.

\paragraph{Symmetric \ttv distance} To make the distance computation robust, we propose to use the cosine distance between normalized embeddings:
\[
d_{\text{sym}}(F_a, F_b) = d_\text{cos} \Big(\frac{F_a}{F_a + F_b}, \frac{F_b}{F_a + F_b}\Big),
\]
where $d_\text{cos}$ is the cosine distance, $F_a$ and $F_b$ are the two task embeddings (\ie, the diagonal of the Fisher Information computed on the same probe network), and the division is element-wise.
This is a symmetric distance which we expect to capture semantic similarity between two tasks. For example, we show in Fig. \ref{fig:taxonomical_distance} that it correlates well with the taxonomical distance between species on  iNaturalist.

On the other hand, precisely for this reason, this distance is ill-suited for tasks such as model selection, where the (intrinsically asymmetric) transfer distance is more relevant.

\paragraph{Asymmetric \ttv distance} In a first approximation, that does not consider either the model or the training procedure used, positive transfer between two tasks depends both on the similarity between two tasks and on the complexity of the first. Indeed, pre-training on a general but complex task such as ImageNet often yields a better result than fine-tuning from a close dataset of comparable complexity. In our case, complexity can be measured as the distance from the trivial embedding. This suggests the following asymmetric score, again improperly called a ``distance'' despite being asymmetric and possibly negative:
\[
d_{\text{asym}}(t_a \to t_b) = d_{\text{sym}}(t_a, t_b) - \alpha d_\text{sym}(t_a, t_0),
\]
where $t_0$ is the trivial embedding, and $\alpha$ is an hyperparameter. This has the effect of bring more complex models closer. The hyper-parameter $\alpha$ can be selected based on the meta-task. In our experiments, we found that the best value of $\alpha$ ($\alpha=0.15$ when using a ResNet-34 pretrained on ImageNet as the probe network) is robust to the choice of meta-tasks.

\section{\textsc{model2vec}: task/model co-embedding}
\label{sec:model_embedding}

By construction, the \ttv distance ignores details of the model and only relies on the task.
If we know what task a model was trained on, we can represent the model by the embedding of that task.  However, in general we may not have such information ({\em e.g.}, black-box models or hand-constructed feature extractors).  We may also have multiple models trained on the same task
with different performance characteristics. To model the joint interaction between task and model (\ie, architecture and training algorithm), we aim to learn a joint embedding of the two.

We consider for concreteness the problem of learning a joint embedding for model selection.
In order to embed models in the task space so that those near a task are likely to perform well on that task, we formulate the following meta-learning problem: Given $k$ models, their \mtv{} embedding are the vectors $m_i = F_i + b_i$,
where $F_i$ is the task embedding of the task used to train model $m_i$ (if available, else we set it to zero), and $b_i$ is a learned ``model bias'' that perturbs the task embedding to account for particularities of the model. We learn $b_i$ by optimizing a $k$-way cross entropy loss to predict the best model given the task distance (see Supplementary Material):
\[
\mathcal{L} = \E[ -\log p(m\,|\,d_\text{asym}(t, m_0), \ldots, d_\text{asym}(t, m_k))].
\]
After training, given a novel query task $t$, we can then predict the best model for it as the $\arg\max_i d_\text{asym}(t, m_i)$, that is, the model $m_i$ embedded closest to the query task.

\begin{figure*}
    \centering
    \includegraphics[height=.38\linewidth]{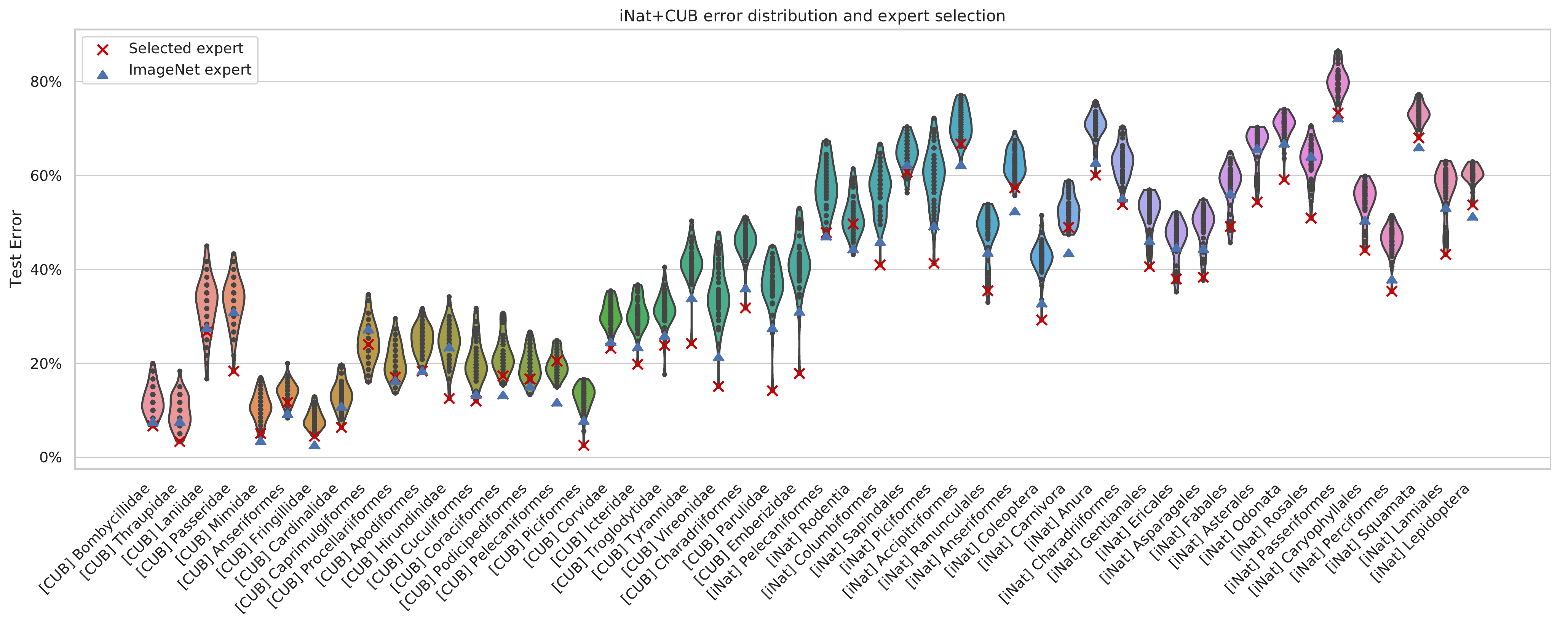}
    \caption{\textbf{\ttv often selects the best available experts.}
    Violin plot of the distribution of the final test error (shaded plot) on tasks from the CUB-200 dataset (columns) obtained by training a linear classifier over several expert feature extractors (points). Most specialized feature extractors perform similarly on a given task, and generally are similar or worse than a generic feature extractor pre-trained on ImageNet (blue triangles). However, in some cases a carefully chosen expert, trained on a relevant task, can greatly outperform all other experts (long whisker of the violin plot). The  model selection algorithm based on \ttv  can, without training, suggest an expert to use for the task (red cross, lower is better). \ttv mostly recover the optimal, or close to optimal, feature extractor to use without having to perform an expensive brute-force search over all possibilities. Columns are ordered by norm of the task embedding: Notice tasks with lower embedding norm have lower error and more ``complex'' task (task with higher embedding norm) tend to benefit more from a specialized expert.
    }
    \label{fig:model_recommendation}
\end{figure*}

\section{Experiments}
\label{sec:experiments}

We test \ttv on a large collection of tasks and models, related to different degrees. Our experiments aim to test both qualitative properties of the embedding and its performance on meta-learning tasks. We use an off-the-shelf ResNet-34 pretrained on ImageNet as our probe network, which we found to give the best overall performance (see Sect. \ref{sec:model-selection}). The collection of tasks is generated starting from the following four main datasets. \textbf{iNaturalist} \cite{van2018inaturalist}: Each task extracted  corresponds to species classification in a given taxonomical order. For instance, the \textit{``Rodentia task''} is to classify  species of rodents. Notice that each task is defined on a separate subset of the images in the original dataset; that is, the domains of the tasks are disjoint. \textbf{CUB-200} \cite{WahCUB_200_2011}: We use the same procedure as iNaturalist to create tasks. In this case, all tasks are classifications inside orders of birds (the \textit{aves} taxonomical class), and have generally much less training samples than corresponding tasks in iNaturalist. \textbf{iMaterialist} \cite{iMatFGVC5} and \textbf{DeepFashion} \cite{liu2016deepfashion}: Each image in both datasets is associated with several binary attributes (\eg, style attributes) and categorical attributes (\eg, color, type of dress, material). We binarize the categorical attributes, and consider each attribute as a separate task. Notice that, in this case, all tasks share the same domain and are naturally correlated.

In total, our collection of tasks has 1460 tasks (207 iNaturalist, 25 CUB, 228 iMaterialist, 1000 DeepFashion).
While a few tasks have many training examples ({\em e.g.}, hundred thousands), most have just hundreds or thousands of samples. This simulates the heavy-tail distribution of data in real-world applications.

Together with the collection of tasks, we collect several ``expert'' feature extractors. These are ResNet-34 models pre-trained on ImageNet and then fine-tuned on a specific task or collection of related tasks (see Supplementary Materials for details). We also consider a ``generic''expert pre-trained on ImageNet without any  finetuning. Finally, for each combination of expert feature extractor and task, we trained a linear classifier on top of the expert in order to solve the selected task using the expert.

In total, we trained 4,100 classifiers, 156 feature extractors and 1,460 embeddings. The total effort to generate the final results was about 1,300 GPU hours.

\paragraph{Meta-tasks.} In Sect. \ref{sec:model-selection}, for a given task we aim to predict, using \ttv, which expert feature extractor will yield the best classification performance. In particular, we formulate two model selection meta-tasks: \textbf{iNat + CUB} and \textbf{Mixed}. The first consists of 50 tasks and experts  from iNaturalist and CUB, and aims to test fine-grained expert selection in a restricted domain. The second contains a mix of 26 curated experts and 50 random tasks extracted from all datasets, and aims to test model selection between different domains and tasks (see Supplementary Material for details).

\subsection{Task Embedding Results}

\paragraph{Task Embedding qualitatively reflects taxonomic distance for iNaturalist}
For tasks extracted from the iNaturalist dataset (classification of species), the taxonomical distance between orders provides a natural metric of the semantic similarity between tasks. In Figure \ref{fig:taxonomical_distance} we compare the symmetric \ttv distance with the taxonomical distance, showing strong agreement.

\paragraph{Task embedding for iMaterialist} In Fig.~\ref{fig:taxonomic_qualitative} we show a t-SNE visualization of the embedding for iMaterialist and iNaturalist tasks. Task embedding yields interpretable results: Tasks that are correlated in the dataset, such as binary classes corresponding to the same categorical attribute, may end up far away from each other and close to other tasks that are semantically more similar (\eg, the \textit{jeans} category task is close to the \textit{ripped} attribute and the \textit{denim} material). This is reflected in the mixture of colors of semantically related nearby tasks, showing non-trivial grouping.

We also compare the \ttv embedding with a domain embedding baseline, which only exploits the  input distribution $p(x)$ rather than the task distribution $p(x,y)$. While some tasks are highly correlated with their domain (\eg, tasks from iNaturalist), other tasks differ only on the labels (\eg, all the attribute tasks of iMaterialist, which  share the same clothes domain). Accordingly, the domain embedding recovers similar clusters on iNaturalist. However, on iMaterialst domain embedding collapses all tasks to a single uninformative cluster (not a single point due to slight noise in embedding computation).

\paragraph{Task Embedding encodes task difficulty}
The scatter-plot in Fig.~\ref{fig:model_recommendation} compares the norm of embedding vectors vs. performance of the best expert (or task specific model for cases where we have the diagonal computed). As shown analytically for the two-layers model, the norm of the task embedding correlates with the complexity of the task also on real tasks and architectures.

\begin{figure}
    \centering
    \includegraphics[width=0.8\columnwidth]{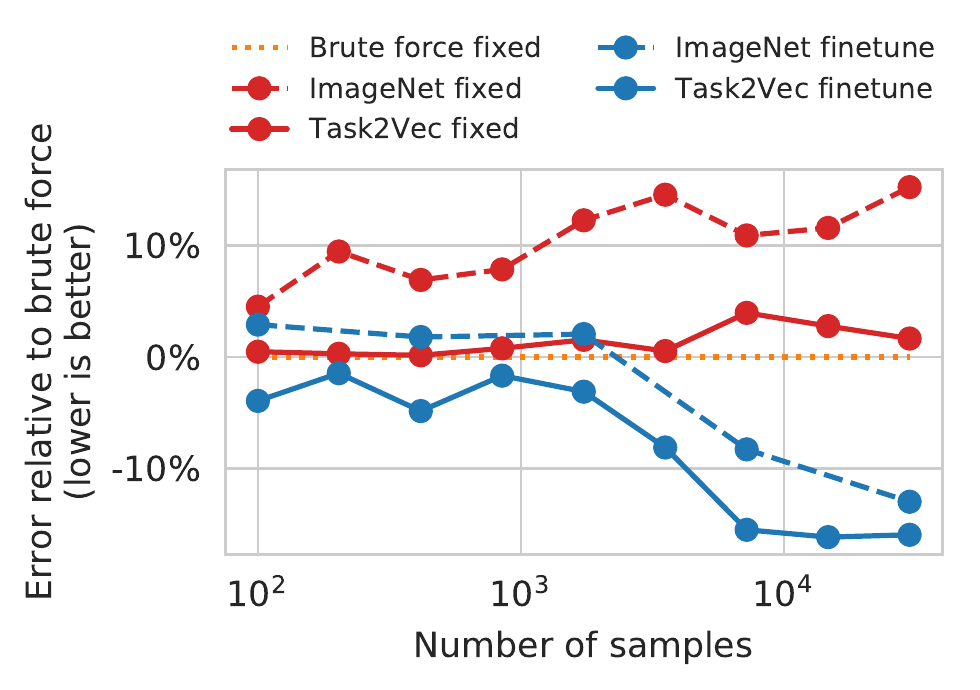}
    \caption{\textbf{\ttv improves results at different dataset sizes and training conditions:} Performance of  model selection on a subset of 4 tasks as a function of the number of samples available to train relative to optimal model selection (dashed orange). Training a classifier on the feature extractor selected by \ttv (solid red) is always better than using a generic ImageNet feature extractor (dashed red). The same holds when allowed to fine-tune the feature extractor (blue curves). Also notice that in the low-data regime fine-tuning the ImageNet feature extractor is more expensive and has a worse performance than accurately selecting a good fixed feature extractor.
    }
    \label{fig:data_efficiency}
\end{figure}

\begin{table}
    \small
    \centering
    \begin{tabular}{c|cc}
        Probe network & Top-10 & All \\
        \hline
        Chance & +13.95\% & +59.52\% \\
        VGG-13 & +4.82\% & +38.03\% \\
        DenseNet-121 & +0.30\% & +10.63\% \\
        ResNet-13 & \textbf{+0.00\%} & \textbf{+9.97\%}
    \end{tabular}
    \caption{ \textbf{Choice of probe network.} Mean relative error increase over the ground-truth optimum on the iNat+CUB meta-task for different choices of the probe-network. We also report the performance on the top 10 tasks with more samples to show how data size affect different architectures.}
    \label{fig:probe_network_choice}
\end{table}

\subsection{Model Selection}
\label{sec:model-selection}

Given a task, our aim is to select an expert feature extractor that maximizes the classification performance on that task. We propose two strategies: (1) embed the task and select the feature extractor trained on the most similar task, and (2) jointly embed the models and tasks, and select a model using the learned metric (see Section \ref{sec:model_embedding}). Notice that (1) does not use knowledge of the model performance on various tasks, which makes it more widely applicable but requires we know what task a model was trained for and may ignore the fact that models trained on slightly different tasks may still provide an overall better feature extractor (for example by over-fitting less to the task they were trained on).

In Table \ref{fig:model_recommendation_table} we compare the overall results of the various proposed metrics on the model selection meta-tasks. On both the iNat+CUB and Mixed meta-tasks, the Asymmetric \ttv model selection is close to the ground-truth optimal, and significantly improves over both chance, and over using an generic ImageNet expert.  Notice that our method has $O(1)$ complexity, while searching over a collection of $N$ experts is $O(N)$.

\paragraph{Error distribution} In Fig.~\ref{fig:model_recommendation} we show in detail the error distribution of the experts on multiple tasks.
It is interesting to notice that the classification error obtained using most experts clusters around some mean value, and little improvement is observed over using a generic expert. On the other hand, a few optimal experts can obtain a largely better performance on the task than a generic expert. This confirms the importance of having access to a large collection of experts when solving a new task, especially if few training data are available. But this collection can only be efficiently exploited if an algorithm is given to efficiently find one of the few experts for the task, which we propose.

\begin{table*}[ht]
    \centering
    \begin{tabular}{c|c|ccccc}
        Meta-task & Optimal & Chance & ImageNet & \ttv & Asymmetric \ttv & \mtv \\
        \hline
        iNat + CUB & 31.24 & +59.52\% & +30.18\% & +42.54\% & +9.97\%  & \textbf{+6.81\%} \\
        Mixed & 22.90 & +112.49\% & +75.73\% & +40.30\% & +29.23\% & \textbf{+27.81\%} \\
    \end{tabular}
    \caption{\textbf{Model selection performance of different metrics.} Average optimal error obtained on two meta-learning tasks by exhaustive search over the best expert, and relative error increase when using cheaper model selection methods. Always picking a fixed good general model (\eg, a model pretrained on ImageNet) performs better than picking an expert at random (chance). However, picking an expert using the Asymmetric \ttv distance can achieve an overall better performance than using a general model. Notice also the improvement over the Symmetric version, especially on iNat + CUB, where experts trained on very similar tasks may be too simple to yield good transfer, and should be avoided. }
    \label{fig:model_recommendation_table}
\end{table*}

\paragraph{Dependence on task dataset size}

Finding experts is especially important when the task we are interested in has relatively few samples. In Fig.~\ref{fig:data_efficiency} we show how the performance of \ttv varies on a model selection task as the number of samples varies. At all sample sizes \ttv is close to the optimum, and improves over selecting a generic expert (ImageNet), both when fine-tuning and when training only a classifier. We observe that the best choice of experts is not affected by the dataset size, and that even with few examples \ttv is able to find the optimal experts.

\paragraph{Choice of probe network}

In Table \ref{fig:probe_network_choice} we show that DenseNet \cite{huang2017densely} and ResNet architectures \cite{he2016deep} perform significantly better when used as probe networks to compute the \ttv embedding than a VGG \cite{simonyan2014very} architecture.

\section{Related Work}
\label{sec:related-work}

\paragraph{Task and Domain embedding.}
Tasks distinguished by their domain can be understood simply in terms of image statistics.
Due to the bias of different datasets, sometimes a benchmark task may be identified just by looking at a few images \cite{torralba2011unbiased}. The question
of determining what summary statistics are useful (analogous to our choice of probe network)
has also been considered, for example \cite{edwards2016towards} train an autoencoder that learns
to extract fixed dimensional summary statistics that can reproduce many different datasets
accurately. However, for general vision tasks which apply to all natural images, the domain
is the same across tasks.

Taskonomy \cite{zamir2018taskonomy} explores the structure of the space of tasks, focusing on the question of effective knowledge transfer in a curated collection of 26 visual tasks, ranging from classification to 3D reconstruction, defined on a common domain. They compute pairwise transfer distances
between pairs of tasks and use the results to compute a directed hierarchy. Introducing novel tasks requires computing the pairwise distance with tasks in the library. In contrast, we focus on a larger library of 1,460 fine-grained classification tasks both on same and different domains, and show that it is possible to represent tasks in a topological space with a constant-time embedding. The large task collection and cheap embedding costs allow us to tackle new meta-learning problems.

\paragraph{Fisher kernels}
Our work takes inspiration from Jaakkola and Hausler \cite{jaakkola1999exploiting}. They propose the ``Fisher Kernel'', which uses the gradients of a generative model score function
as a representation of similarity between data items
\[
K(x^{(1)}, x^{(2)}) = \nabla_\theta \log P (x^{(1)}|\theta)^{T} F^{-1}\nabla_\theta \log P (x^{(2)}|\theta).
\]
Here $P(x | \theta)$ is a parameterized generative model and $F$ is the Fisher information matrix.
This provides a way to utilize generative
models in the context of discriminative learning. Variants of the Fisher kernel have found wide use
as a representation of images~\cite{perronnin2010improving,sanchez2013image}, and other structured data such as protein molecules~\cite{jaakkola1999using} and text~\cite{saunders2003string}. Since the generative model can be learned on unlabelled data, several works have investigated the use of Fisher kernel for unsupervised learning~\cite{holub2005combining,seeger2000learning}.
\cite{van2011learning} learns a metric on the
Fisher kernel representation similar to our metric learning approach. Our approach differs in
that we use the FIM as a representation of a whole dataset (task) rather than using model gradients
as representations of individual data items.

\paragraph{Fisher Information for CNNs}
Our approach to task embedding makes use of the Fisher Information matrix of a neural network
as a characterization of the task.  Use of Fisher information for neural networks was popularized
by Amari \cite{amari1998natural} who advocated optimization using natural gradient descent
which leverages the fact that the FIM is an appropriate parameterization-independent metric on
statistical models. Recent work has focused on approximates of FIM appropriate in this
setting (see e.g., \cite{heskes2000natural,finn2017model,martens2015optimizing}).  FIM has also been proposed
for various regularization schemes \cite{achille2018emergence,arora2018stronger,liang2017fisher,mroueh2017fisher}, analyze learning dynamics of deep networks \cite{achille2017critical}, and to overcome catastrophic forgetting
\cite{kirkpatrick2017overcoming}.

\paragraph{Meta-learning and Model Selection}
The general problem of meta-learning has a long history with much recent work dedicated to
problems such as neural architecture search and hyper-parameter estimation.  Closely
related to our problem is work on selecting from a library of classifiers to solve a
new task~\cite{smith2014recommending,abdulrahman2018speeding,leite2012selecting}.  Unlike
our approach, these usually address the question via land-marking or active testing, in
which a few different models are evaluated and performance of the remainder estimated by
extension. This can be viewed as a problem of completing a matrix defined by performance
of each model on each task.

A similar approach has been taken in computer vision for selecting a detector for a new
category out of a large library of detectors~\cite{matikainen2012model,zhang2014predicting,wang2015model}.

\section{Discussion}

\ttv is an efficient way to represent a task, or the corresponding dataset, as a fixed dimensional vector. It has several appealing properties, in particular its norm correlates with the test error obtained on the task, and the cosine distance between embeddings correlates with natural distances between tasks, when available, such as the taxonomic distance for species classification, and the fine-tuning distance for transfer learning. Having a representation of tasks paves the way for a wide variety of meta-learning tasks. In this work, we focused on selection of an expert feature extractor in order to solve a new task, especially when little training data is present, and showed that using \ttv to select an expert from a collection can sensibly improve test performance while adding only a small overhead to the training process.

Meta-learning on the space of tasks is an important step toward general artificial intelligence. In this work, we introduce a way of dealing with thousands of tasks, enough to enable reconstruct a topology on the task space, and to test meta-learning solutions. The current experiments highlight the usefulness of our methods. Even so, our collection does not capture the full complexity and variety of tasks that one may encounter in real-world situations. Future work should further test effectiveness, robustness, and limitations of the embedding on larger and more diverse collections.

{\small
\bibliographystyle{ieee}
\bibliography{references}
}

\appendix
\renewcommand{\DDX}[1]{\frac{\partial}{\partial{#1}}}
\renewcommand{\DXDY}[2]{\frac{\partial{#1}}{\partial{#2}}}
\renewcommand{\HXY}[2]{\frac{\partial^2\ell}{\partial{#1}\partial{#2}}}
\renewcommand{\FXY}[2]{\frac{\partial \ell}{\partial{#1}}\left(\frac{\partial \ell}{\partial{#2}}\right)^T}

\onecolumn

\section{Analytic FIM for two-layer model}
\label{sec:understand_fisher}

Assume we have data points $(x_i, y_i), i=1 \ldots n$ and $y_i \in \{0,1\}$. Assume that a fixed feature extractor
applied to data points $x$ yields features $z = \phi(x) \in \mathbb{R}^d$ and a linear model with parameters $w$ is trained
to model the conditional distribution $p_i = P(y_i = 1| x_i) = \sigma\left(w^T\phi(x_i)\right),$ where
$\sigma$ is the sigmoid function. The gradient of the cross-entropy loss with respect to the linear model
parameters is:
\[
\DXDY{\ell}{w} = \frac{1}{N}\sum_i (y_i - p_i)\phi(x_i),
\]
and the empirical estimate of the Fisher information matrix is:
\begin{align*}
F &= \E \Big[\FXY{w}{w}\Big] = \E_{y \sim p_w(y|x)} \frac{1}{N} \sum_i \phi(x_i)(y_i - p_i)^2\phi(x_i)^T\\
&= \frac{1}{n} \sum_i \phi(x_i)(1 - p_i)p_i\phi(x_i)^T
\end{align*}

In general, we are also interested in the Fisher information of the parameters of
the feature extractor $\phi(x)$ since this is independent of the specifics of the
output space $y$ (e.g., for k-way classification).  Consider a 2-layer network
where the feature extractor uses a sigmoid non-linearity:
\begin{eqnarray*}
p = \sigma(w^T z) & z_k = \sigma(U_k^T x)
\end{eqnarray*}
and the matrix $U$ specifies the feature extractor parameters and $w$ are parameters
of the task-specific classifier. Taking the gradient w.r.t. parameters we have:
\begin{align*}
\DXDY{\ell}{w_j} &= (y-p)z_j \\
\DXDY{\ell}{U_{kj}} &= (y-p) w_k z_k (1-z_k) x_j
\end{align*}
The Fisher Information Matrix (FIM) consists of blocks:
\begin{align*}
\FXY{w_i}{w_j} &= (y-p)^2 z_i z_j \\
\FXY{U_{ki}}{w_{j}} &= (y-p)^2 z_j z_k (1-z_k) x_i \\
\FXY{U_{li}}{U_{kj}} &= (y-p)^2 w_k z_k (1-z_k) w_l z_l (1-z_l) x_i x_j
\end{align*}

We focus on the FIM of the probe network parameters which is
independent of the dimensionality of the output layer and write it
in matrix form as:
\[
\FXY{U_{l}}{U_{k}} = (y-p)^2 (1-z_k) z_k (1-z_l) z_l w_k w_l xx^T
\]
Note that each block $\{l,k\}$ consists of the same matrix $(y-p)^2 \cdot xx^T$
multiplied by a scalar $S_{kl}$ given as:
\[
S_{kl} = (1-z_k) z_k (1-z_l) z_l w_k w_l
\]
We can thus write the whole FIM as the expectation of a Kronecker product:
\[
F = \E [(y-p)^2 \cdot S \otimes xx^T]
\]
where the matrix $S$ can be written as
\[
S =  ww^T \odot zz^T \odot (1-z)(1-z)^T
\]

Given a task described by N training samples $\{(x_e,y_e)\}$, the FIM can
be estimated empirically as
\begin{align*}
F &= \frac{1}{N} \sum_e p_e (1-p_e) \cdot S_e \otimes x_e x_e^T \\
S_e &=  ww^T \odot z_e z_e^T \odot (1-z_e)(1-z_e)^T
\end{align*}
where we take expectation over $y$ w.r.t. the predictive distribution
$y \sim p_w(y|x)$.

\begin{figure*}
    \centering
    \begin{tabular}{cc}
    \includegraphics[width=0.3\linewidth]{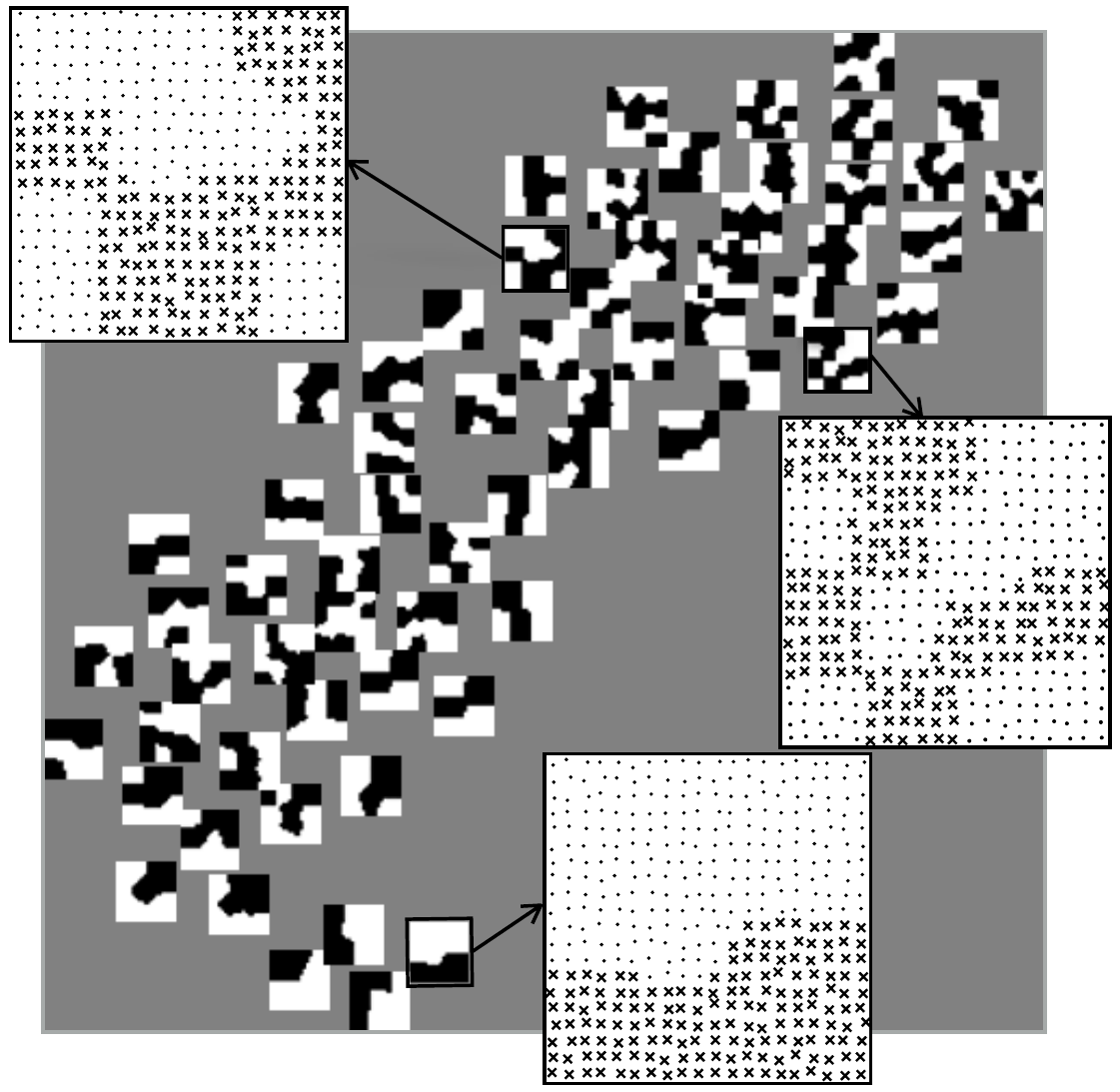} &
    \includegraphics[width=0.3\linewidth]{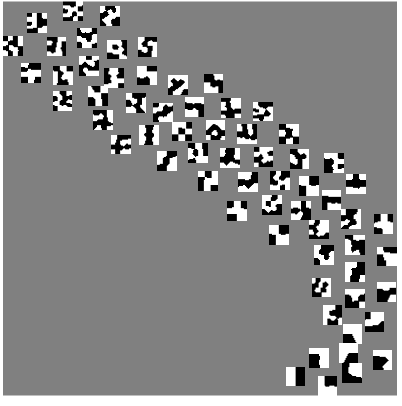} \\
    (a) Random linear + ReLU &  (b) Polynomial of degree three
    \end{tabular}
    \caption{Task embeddings computed for a probe network consisting of (a) 10 random linear + ReLU features and (b) degree three polynomial features projected to 2D using t-SNE. The tasks are random binary partitions of the unit square visualized in each icon (three tasks are visualized on the left) and cannot be distinguished based purely on the input domain without considering target labels. Note that qualitatively similar tasks group together, with more complex tasks (requiring complicated decision boundaries) separated from simpler tasks.}
    \label{fig:toy-task-embedding}
\end{figure*}

\paragraph{Example toy task embedding}
As noted in the main text, the FIM depends on the domain embedding, the particular task and its complexity.
We illustrate these properties of the task embedding using an ``toy'' task space
illustrated in Figure~\ref{fig:toy-task-embedding}. We generate 64 binary classification tasks by clustering a uniform grid of points in the XY plane into $k \in [3,16]$ clusters using $k$-means
and assigning a half of them to one category.  We consider two different feature
extractors, which play the role of ``probe network".  One is a collection of polynomial functions of degree $d=3$, the second
is $10$ random linear features of the form $\max(0, ax + by + c)$ where $a$ and $b$ are
sampled uniformly between $[-1/2, 1/2]$ and $c$ between $[-1, 1]$.

\section{Robust Fisher Computation}

Consider again the loss function (parametrized with the covariance matrix $\Sigma$ instead of the precision matrix $\Lambda$ for convenience of notation):
\[
    L(\hat{w}; \Sigma) = \E_{w\sim \N(\hat{w},\Sigma)} [H_{p_w,\hat{p}}(y|x)]
+ \beta \KL{\N(\hat{w}, \Sigma)}{\N(0, \sigma^2 I)}.
\]
We will make use of the fact that the Fisher Information matrix is a positive semidefinite approximation of the Hessian $H$ of the cross-entropy loss, and coincide with it in local minima \cite{martens14new}.  Expanding to the second order around $\hat{w}$, we have:
\begin{align*}
    L(\hat{w}; \Sigma) =& \E_{w\sim \N(\hat{w},\Sigma)} [H_{p_{\hat{w}},\hat{p}}(y|x) + \nabla_w H_{p_{\hat{w}},\hat{p}}(y|x) (w - \hat{w}) + \frac{1}{2}(w-\hat{w})^T H (w-\hat{w})]  + \beta \KL{\N(\hat{w}, \Sigma)}{\N(0, \sigma^2 I)}\\
=& H_{p_{\hat{w}},\hat{p}}(y|x) + \frac{1}{2} \tr(\Sigma H) + \beta \KL{\N(\hat{w}, \Sigma)}{\N(0, \sigma^2 I)}\\
=& H_{p_{\hat{w}},\hat{p}}(y|x) + \frac{1}{2} \tr(\Sigma H) + \frac{\beta}{2} [\frac{\hat{w}^2}{\sigma^2} + \frac{1}{\sigma^2} \tr{\Sigma} + k \log{\sigma^2} - \log(|\Sigma|) - k]
\end{align*}
where in the last line used the known expression for the KL divergence of two Gaussian.
Taking the derivative with respect to $\Sigma$ and setting it to zero, we obtain that the expression loss is minimized when
$
\Sigma^{-1} = \frac{2}{\beta} \Big(H + \frac{\beta}{2\sigma^2} I\Big),
$
or, rewritten in term of the precision matrices, when
\[
\Lambda = \frac{2}{\beta} \Big(H + \frac{\beta \lambda^2}{2} I\Big),
\]
where we have introduced the precision matrices $\Lambda = \Sigma^{-1}$ and  $\lambda^2 I =1/\sigma^2 I$.

We can then obtain an estimate of the Hessian $H$ of the cross-entropy loss at the point $\hat{w}$, and hence of the FIM, by minimizing the loss $L(\hat{w}, \Lambda)$ with respect to $\Lambda$. This is a more robust approximation than the standard definition, as it depends on the loss in a whole neighborhood of $\hat{w}$ of size $\propto  \Lambda$, rather than from the derivatives of the loss at a point.
To further make the estimation more robust, and to reduce the number of parameters, we constrain $\Lambda$ to be diagonal, and constrain weights $w_{ij}$ belonging to the same filter to have the same precision $\Lambda_{ij}$.
Optimization of this loss can be performed easily using Stochastic Gradient Variational Bayes, and in particular using the local reparametrization trick of \cite{kingma2015variational}.

The prior precision $\lambda^2$ should be picked according to the scale of the weights of each layer. In practice, since the weights of each layer have a different scale, we found it useful to select a different $\lambda^2$ for each layer, and train it together with $\Lambda$,

\section{Details of the experiments}

\subsection{Training of experts and classifiers}

Given a task, we train an \textit{expert} on it by fine-tuning an off-the-shelf ResNet-34 pretrained on ImageNet%
\footnote{\url{https://pytorch.org/docs/stable/torchvision/models.html}}. Fine-tuning is performed by first fixing the weights of the network and retraining from scratch only the final classifier for 10 epochs using Adam, and then fine-tuning all the network together with SGD for 60 epochs with weight decay 5e-4, starting from learning rate 0.001 and  decreasing it by a factor 0.1 at epochs 40.

Given an expert, we train a classifier on top of it by replacing the final classification layer and training it with Adam for 16 epochs. We use weight decay 5e-4 and learning rate 1e-4.

The tasks we train on generally have different number of samples and unbalanced classes. To limit the impact of this imbalance on the training procedure, regardless of the total size of the dataset, in each epoch we always sample 10,000 images with replacement, uniformly between classes. In this way, all epochs have the same length and see approximately the same number of examples for each class. We use this balanced sampling in all experiments, unless noted otherwise.

\subsection{Computation of the \ttv embedding}

As the described in the main text, the \ttv embedding is obtained by choosing a probe network, retraining the final classifier on the given task, and then computing the Fisher Information Matrix for the weights of the probe network.

Unless specified otherwise, we use an off-the-shelf ResNet-34 pretrained on ImageNet as the probe network.
The Fisher Information Matrix is computed in a robust way minimizing the loss function $L(\hat{w}; \Lambda)$ with respect to the precision matrix $\Lambda$, as described before.
To make computation of the embedding faster, instead of waiting for the convergence of the classifier, we train the final classifier for 2 epochs using Adam and then we continue to train it jointly with the precision matrix $\Lambda$ using the loss $L(\hat{w}; \Lambda)$. We constrain $\Lambda$ to be positive by parametrizing it as $\Lambda = \exp(L)$, for some unconstrained variable $L$. While for the classifier we use a low learning rate (1e-4), we found it useful to use an higher learning rate (1e-2) to train $L$.

\subsection{Training the \mtv embedding}

As described in the main text, in the \mtv embedding we aim to learn a vector representation $m_j = F_j + b_j$ of the $j$-th model in the collection, which represents both the task the model was trained on (through the \ttv embedding $F_j$), and the particularities of the model (through the learned parameter $b_j$).

We learn $b_j$ by minimizing a $k$-way classification loss which, given a task $t$, aims to select the model that performs best on the task among a collection of $k$ models. Multiple models may perform similarly and close to optimal: to preserve this information, instead of using a one-hot encoding for the best model, we train using soft-labels obtained as follows:
\[
\hat{p}(y_i) = \text{Softmax}\Big(- \alpha \frac{\text{error}_i - \text{mean}(\text{error}_i)}{\text{std}(\text{error}_{i})}\Big),
\]
where $\text{error}_{i,j}$ is the ground-truth test error obtained by training a classifier for task $i$ on top of the $j$-th model. Notice that for $\alpha \gg 1$, the soft-label $y^j_i$ reduces to the one-hot encoding of the index of the best performing model. However, for lower $\alpha$'s, the vector $y_i$ contains richer information about the relative performance of the models.

We obtain our prediction in a similar way: Let $d_{i, j} = d_\text{asym}(t_i, m_j)$, then we set our model prediction to be
\[
p(y|d_{i,0},\ldots,d_{i,k}) = \text{Softmax}(-\gamma\, d_i),
\]
where the scalar $\gamma > 0$ is a learned parameter. Finally, we learn both the $m_j$'s and $\gamma$ using a cross-entropy loss:
\[
\mathcal{L} = \frac{1}{N} \sum_{i=0}^N \E_{y_i \sim \hat{p}}[p(y_i|d_{i,0},\ldots,d_{i,k})],
\]
which is minimized precisely when $p(y|d_{i,0},\ldots,d_{i,k}) = \hat{p}(y_i)$.

In our experiments we set $\alpha=20$, and  minimize the loss using Adam with learning rate $0.05$, weight decay $0.0005$, and early stopping after 81 epochs, and report the leave-one-out error (that is, for each task we train using the ground truth of all other tasks and test on that task alone, and report the average of the test errors obtained in this way).

\section{Datasets, tasks and meta-tasks}

Our two model selection meta-tasks, \textbf{iNat+CUB} and \textbf{Mixed}, are curated as follows. For \textbf{iNat+CUB}, we generated 50 tasks and (the same) experts from iNaturalist and CUB. The 50 tasks consist of 25 iNaturalist tasks and 25 CUB tasks to provide a balanced mix from two datasets of the same domain. We generated the 25 iNaturalist tasks by grouping species into orders and then choosing the top 25 orders with the most samples. The number of samples for tasks shows the heavy-tail distribution typical of real data, with the top task having 64,100 samples (the \textit{Passeriformes} order classification task), while most tasks have around $6,000$ samples.

The 25 CUB tasks were similarly generated with 10 order tasks but additionally has 15 Passeriformes family tasks: After grouping CUB into orders, we determined 11 usable order tasks (the only unusable order task, Gaviiformes, has only one species so it makes no sense to train on it). However, one of the orders---Passeriformes---dominated all other orders with 134 species when compared to 3-24 species of other orders. Therefore, we decided to further subdivide the Passeriformes order task into family tasks (\textit{i.e.}, grouping species into families) to provide a more balanced partition. This resulted in 15 usable family tasks (\textit{i.e.}, has more than one species) out of 22 family tasks. Unlike iNaturalist, tasks from CUB have only a few hundreds of samples and hence benefit more from carefully selecting an expert.

In the iNAT+CUB meta-task the classification tasks are the same tasks used to train the experts. To avoid trivial  solutions (always selecting the expert trained on the task we are trying to solve) we test in a leave-one-out fashion: given a classficication task, we aim to select the best expert that was not trained on the same data.

For the \textbf{Mixed} meta-task, we chose 40 random tasks and 25 curated experts from all datasets. The 25 experts were generated from iNaturalist, iMaterialist and DeepFashion (CUB, having fewer samples than iNaturalist, is more appropriate as tasks). For iNaturalist, we trained 15 experts: 8 order tasks and 7 class tasks (species ordered by class), both with number of samples greater than 10,000. For DeepFashion, we trained 3 category experts (upper-body, lower-body, full-body). For iMaterialist, we trained 2 category experts (pants, shoes) and 5 multi-label experts by grouping attributes (color, gender, neckline, sleeve, style). For the purposes of clustering attributes into larger groups for training experts (and color coding the dots in Figure 1), we obtained a de-anonymized list of the iMaterialist Fashion attribute names from the FGCV contest organizers.

The 40 random tasks were generated as follows. In order to balance tasks among all datasets, we selected 5 CUB, 15 iNaturalist, 15 iMaterialist and 5 DeepFashion tasks. Within those datasets, we randomly pick tasks with a sufficient number of validation samples and maximum variety. For the iNaturalist tasks, we group the order tasks into class tasks, filter out the number of validation samples less than 100 and randomly pick order tasks within each class. For the iMaterialist tasks, we similarly group the tasks (e.g. category, style, pattern), filter out tasks with less than 1,000 validation samples and randomly pick tasks within each group. For CUB, we randomly select 2 order tasks and 3 Passeriformes family tasks, and for DeepFashion, we randomly select the tasks uniformly. All this ensures that we have a balanced variety of tasks.

For the \textbf{data efficiency} experiment, we trained on  a subset of the tasks  and experts in the Mixed meta-task: We picked the Accipitriformes, Asparagales, Upper-body, Short Sleeves for the tasks, and the Color, Lepidoptera, Upper-body, Passeriformes, Asterales for the experts. Tasks where selected among those that have more than 30,000 training samples in order to represent all datasets. The experts were also selected to be representative of all datasets, and contain both strong and very weak experts (such as the Color expert).

\pagebreak

\section{Error matrices}

\begin{figure*}[h!]
    \centering
    \includegraphics[height=.57\linewidth]{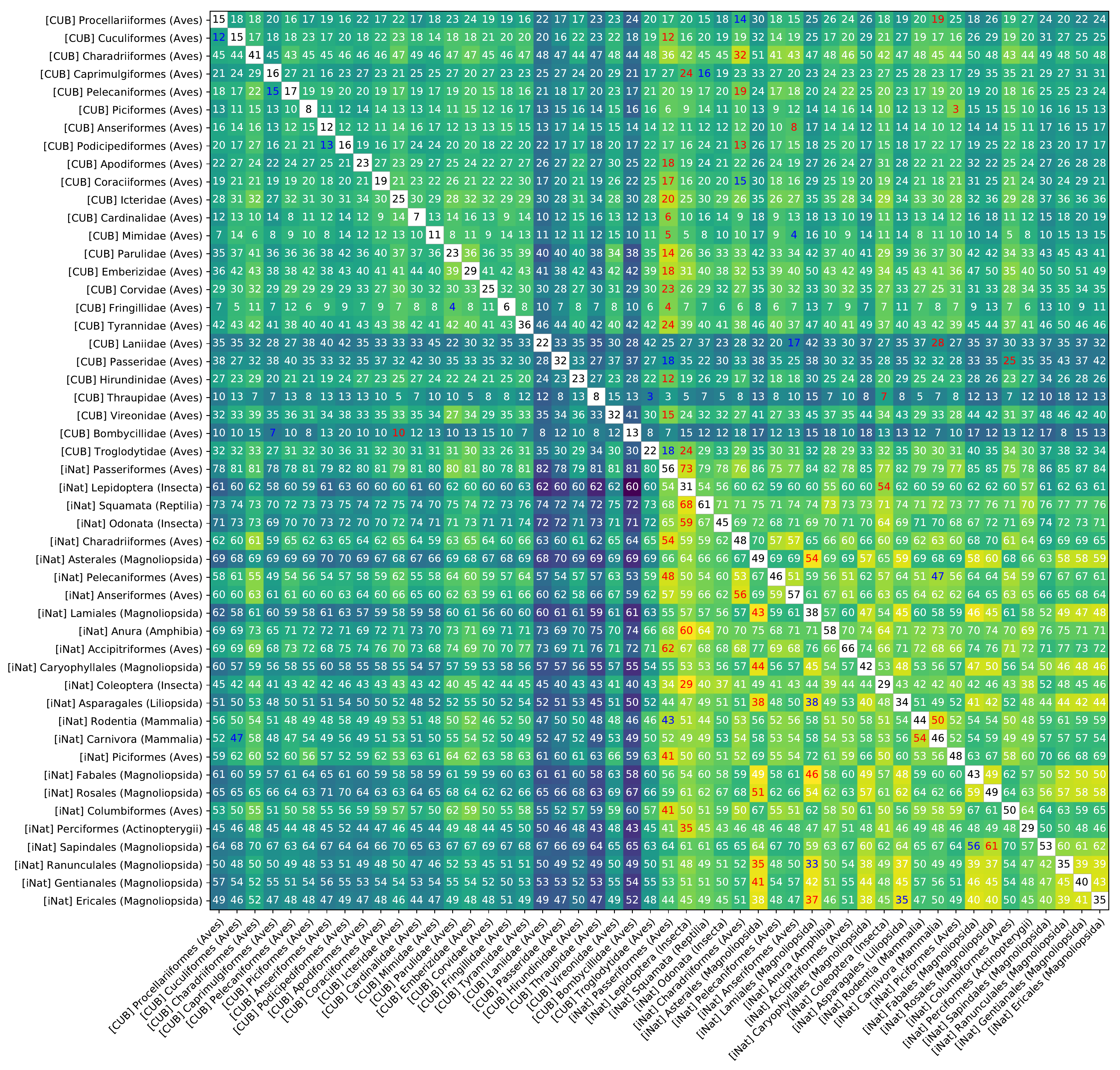}
    \includegraphics[height=.57\linewidth]{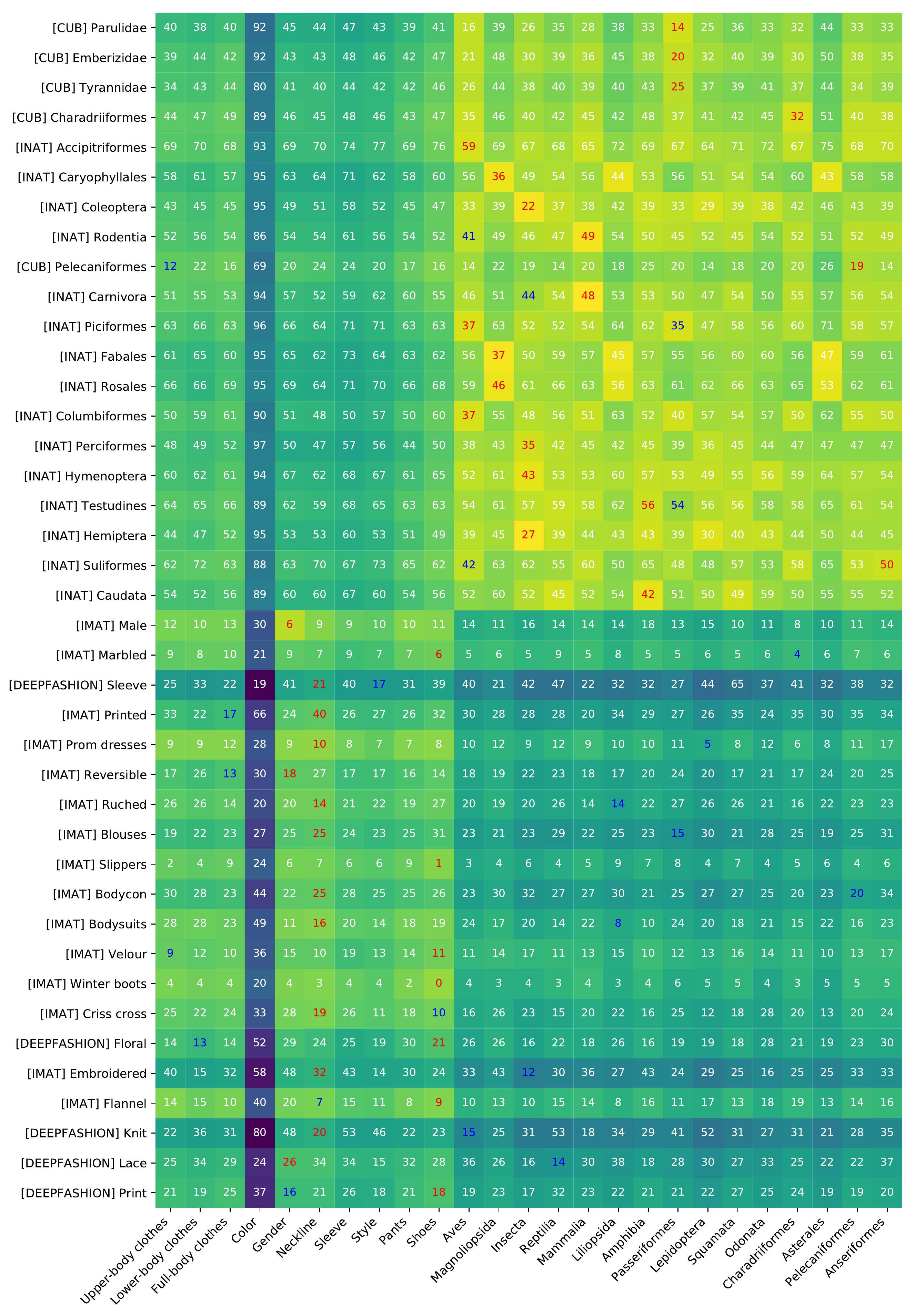}
    \caption{\textbf{Meta-tasks ground-truth error matrices.} (Best viewed magnified). \textbf{(Left)} Error matrix for the CUB+iNat meta-task. The numbers in each cell is the test error obtained by training a classifier on a given combination of task (rows) and expert (columns). The background color represent the Asymmetric \ttv distance between the target task and the task used to train the expert. Numbers in red indicate the selection made by the model selection algorithm based on the Asymmetric \ttv embedding. The (out-of-diagonal) optimal expert (when different from the one selected by our algorithm), is highlighted in blue. \textbf{(Right)} Same as before, but for the Mixed meta-task. }
    \label{fig:inaterror}
\end{figure*}

\end{document}